\documentclass[journal=acsami,manuscript=article]{achemso}

\usepackage[version=3]{mhchem}
\usepackage{multirow}
\usepackage{makecell}
\usepackage{booktabs}
\usepackage{amsmath}
\usepackage{amssymb}
\usepackage{graphicx}
\usepackage{algorithm}
\usepackage{algpseudocode}
\usepackage{hyperref}
\usepackage{float}
\usepackage{xcolor}

\author{Alexander Zhao}
\affiliation{Department of Materials Science and Engineering, Carnegie Mellon University, 5000 Forbes Avenue, Pittsburgh, PA 15213, USA}

\author{Achuth Chandrasekhar}
\affiliation{Department of Mechanical Engineering, Carnegie Mellon University, 5000 Forbes Avenue, Pittsburgh, PA 15213, USA}

\author{Amir Barati Farimani}
\email{barati@cmu.edu}
\affiliation{Department of Mechanical Engineering, Carnegie Mellon University, 5000 Forbes Avenue, Pittsburgh, PA 15213, USA}

\title{PolyJarvis: An LLM-Orchestrated Agent for Automated All-Atom Molecular Dynamics of Amorphous Homopolymers}


\begin{document}

\begin{abstract}
All-atom molecular dynamics (MD) simulations can predict polymer properties from molecular structure, yet their execution requires specialized expertise in force field selection, system construction, equilibration, and property extraction.
We present PolyJarvis, a platform in which a planning agent produces a validated run plan that deterministic stage scripts execute through established simulation toolkits, Enhanced Monte Carlo (EMC) for system construction and LAMMPS for molecular dynamics, exposed as Model Context Protocol (MCP) servers, with a recovery agent consulted only on structured failures and within a fixed decision budget.
Given a repeat-unit SMILES string and target properties, PolyJarvis constructs the amorphous cell, equilibrates it under a mechanized convergence gate, and computes target properties.
Validation is conducted on seven amorphous homopolymers, each run as three replicates that share a protocol frozen per system and use independent random seeds, namely polyethylene (PE), atactic polystyrene (aPS), syndiotactic poly(vinyl chloride) (sPVC), poly(L-lactic acid) (PLLA), poly(ethylene glycol) (PEG), poly(ether ether ketone) (PEEK), and polysulfone (PSU).

Against experimental references, 13 of 19 graded comparisons meet the acceptance criteria (density 5 of 7, glass transition 4 of 7, bulk modulus 4 of 5). The failures are concentrated in the PCFF systems: under-density of aPS and PEG, overestimated glass transitions of the stiff PLLA and PEEK backbones, and an overstiff PEG bulk modulus.
\end{abstract}

\section{Introduction}

Computational prediction of polymer properties such as $T_g$, density, and mechanical moduli is essential for accelerating materials discovery.\cite{bicerano2002prediction}
All-atom MD simulations provide a first-principles route to these predictions,\cite{kremer2009multiscale} but their practical adoption is hindered by the deep expertise required across force field selection, system construction, multi-stage equilibration, and property-specific analysis.\cite{larsen2011molecular, hayashi2022radonpy}
The manual, decision-intensive nature of these workflows limits reproducibility, since analyst choices and finite-ensemble sampling can produce divergent property estimates.\cite{patrone2016uncertainty, suter2025ensemble}

Significant progress has been made in automating the polymer simulation pipeline.
RadonPy\cite{hayashi2022radonpy} provides an end-to-end framework calculating 15 properties for more than 1000 amorphous polymers. SPACIER\cite{nanjo2025spacier} adds Bayesian optimization, Polymatic\cite{abbott2013polymatic} and Polyply\cite{grunewald2022polyply} automate specific MD stages, and PolyArena\cite{polyarena2025} provides standardized evaluation.
Machine learning approaches offer rapid prediction but sacrifice physical interpretability and require extensive training data\cite{tao2021benchmarking, chaudhari2024alloybert}. However, existing tools execute predefined protocols without adapting to the specific polymer system.

Agents based on LLMs have opened new possibilities for scientific automation.
For example, ChemCrow\cite{bran2024chemcrow} integrated chemistry tools with GPT-4 for organic synthesis, DynaMate\cite{guilbert2025dynamate} introduced self-correcting protein--ligand MD, and an autonomous simulation agent was explored for polymer conformations.\cite{liu2025asa}
In materials science specifically, modular LLM agents have been applied to multi-task computational workflows,\cite{chaudhari2025modular} catalyst discovery,\cite{ock2026adsorb} materials knowledge navigation,\cite{zhuang2026natural} polymer design,\cite{nigam2026polymeragent} and biomolecular MD automation.\cite{chandrasekhar2025namdagent}
However, few have demonstrated an LLM agent executing validated all-atom polymer MD simulations with quantitative experimental benchmarking (Table~\ref{tab:comparison_agents}).

\begin{table}[ht]
\centering
\caption{Comparison of LLM-based MD automation frameworks.}
\label{tab:comparison_agents}
\small
\begin{tabular}{@{}llll@{}}
\toprule
\textbf{System} & \textbf{Domain} & \textbf{Validation} & \textbf{Autonomy} \\
\midrule
ChemCrow\cite{bran2024chemcrow} & Organic synth. & Expert eval. & Full \\
DynaMate\cite{guilbert2025dynamate} & Protein--ligand & Task completion & Full \\
MDAgent\cite{shi2025mdagent} & Materials MD & Expert eval. & Semi-auto \\
ASA\cite{liu2025asa} & Polymer conf. & Chain stats & Partial \\
Polymer-Agent\cite{nigam2026polymeragent} & Polymer design & Expert eval. & Full \\
NAMD-Agent\cite{chandrasekhar2025namdagent} & Protein MD & Thermo.\ params & Full \\
\textbf{PolyJarvis} & \textbf{Polymer MD} & \textbf{Exp.\ bench.} & \textbf{Planning, bounded recovery}\\
\bottomrule
\end{tabular}
\par\smallskip
\end{table}

PolyJarvis differs from these frameworks in domain, pipeline coverage, and validation basis. It targets amorphous polymer MD, where system construction and equilibration are the hard steps, rather than small-molecule or biomolecular MD, and it spans the full pipeline from repeat-unit construction to property extraction rather than a single stage. For validation, we compare predicted $T_g$, density, and bulk modulus against experimental references, where prior agents report task completion or rely on expert evaluation.
\section{Methods}

\subsection{System Architecture}

\begin{figure}
\centering
\includegraphics[width=\textwidth]{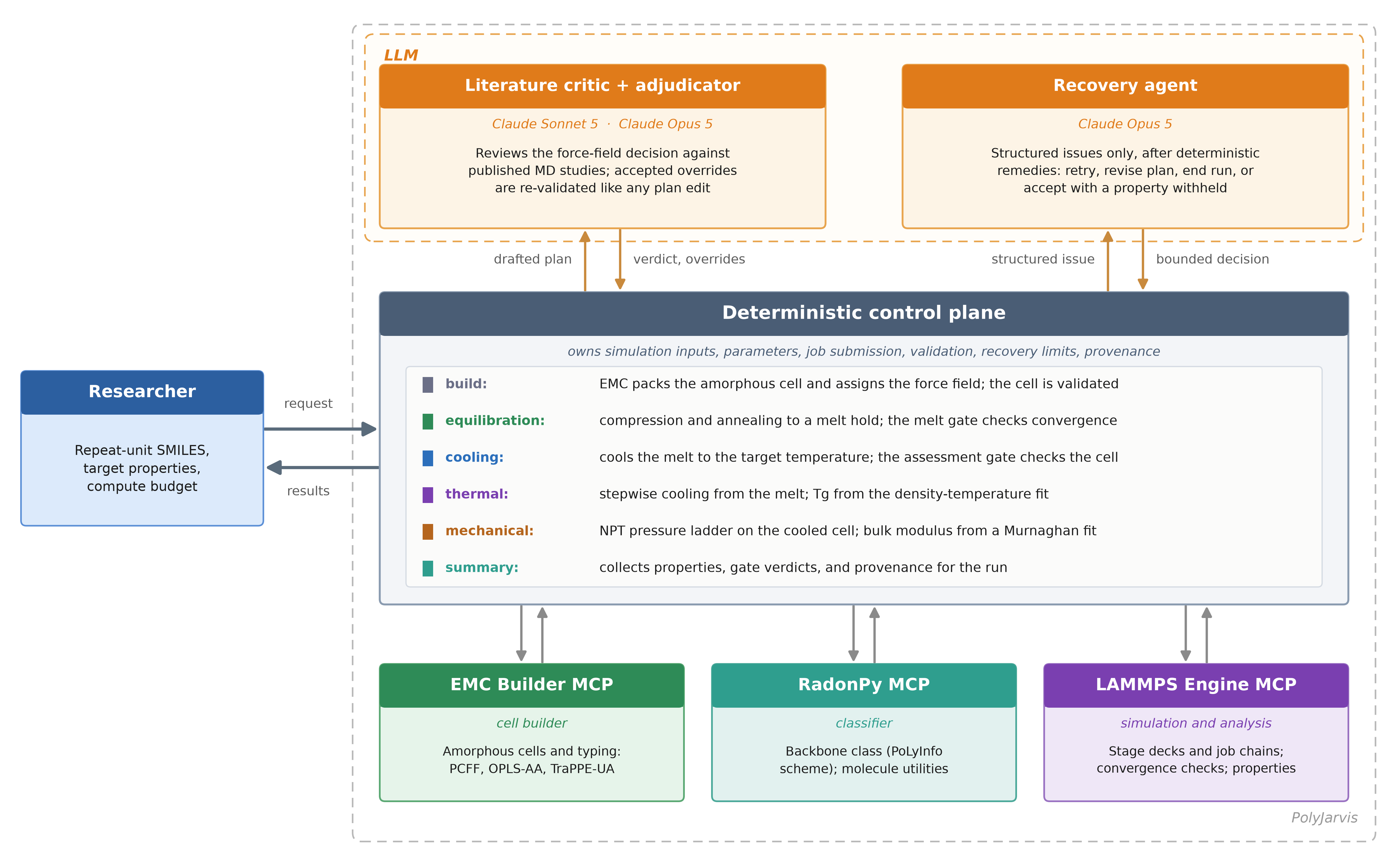}
\caption{PolyJarvis architecture. The deterministic control plane classifies the repeat unit, writes and validates the run plan, and executes the stage scripts, with convergence gates on the melt and on the cell cooled to the target temperature. Language-model calls (orange) are confined to a literature critic with an adjudicator, whose accepted overrides are re-validated, and a recovery agent consulted only on structured issues within a fixed decision budget. All simulation and analysis run through three MCP servers.}
\label{fig:architecture}
\end{figure}

PolyJarvis separates scientific planning from execution (Figure~\ref{fig:architecture}). A planning step turns a repeat-unit SMILES and a set of target properties into a run plan, which is validated structurally and against policy before any simulation is launched. Deterministic stage scripts then execute the plan in a fixed order of cell construction, equilibration, cooling to the target temperature, and the thermal, mechanical, and summary stages. Run state is held on disk per stage and per attempt, so a run resumes from its last accepted stage. The tool layer is three MCP\cite{anthropic2024mcp} servers, an EMC\cite{intveld2003emc} server that builds amorphous cells, a LAMMPS\cite{thompson2022lammps} server for simulation and analysis, and a molecule-builder server based on RadonPy.\cite{hayashi2022radonpy}

The language model (Claude\cite{anthropic2024claude}) enters at two points. In planning, a literature critic reviews the drafted plan against published MD studies of the same polymer and may propose overrides, which an adjudicator accepts or declines. In execution, a recovery agent is consulted only when validation or a stage script returns a structured issue that no deterministic remedy addresses. It may retry, wait and retry, revise the plan, end the run, or accept a property stage with that property withheld. Plan revisions must pass the same bounds as the plan, and the agent has a fixed number of decisions per run and per stage, after which the run ends with its terminating cause recorded (SI Section~S1).

\subsection{Planning and Force-Field Selection}

The repeat unit is assigned one of 21 backbone classes of the PoLyInfo scheme\cite{otsuka2011polyinfo} by RadonPy's classifier, and the class supplies a force-field prior and per-class protocol constants. A functional-group profile of the whole repeat unit is checked against these constants, and disagreements are passed to the critic as advisories. A screen of the repeat unit's chemical moieties flags chemistries known to defeat atom typing, and a flagged repeat unit is assigned a field only after a trial EMC build confirms that the field types it.

\subsection{System Construction}

Each cell contains 10 chains built by EMC at low packing density, with a degree of polymerization that gives chains of about 5~kg\,mol$^{-1}$ and cells of 4{,}940 to 8{,}000 atoms. All MD uses Nos\'{e}--Hoover temperature and pressure control with damping constants of 100 and 1000~fs and a 1~fs timestep, or 2~fs for united-atom models. The executed protocol is listed per system in SI Section~S2.

\subsection{Equilibration and Convergence Gate}
\label{sec:gate}

Equilibration proceeds through energy minimization, an NVT warm-up, NPT densification at high pressure and decompression to 1~atm, NVT annealing above the melt temperature, a ramp to the melt temperature, and NVT and NPT melt holds. The cell is then cooled in NPT blocks to the target temperature and held there. The final melt hold and the final hold at the target temperature are each graded by a mechanized convergence gate, and the graded cell at the target temperature supplies density and the starting cell for the mechanical stage.

The gate evaluates thermodynamic stationarity, chain relaxation, and structure. Stationarity covers density and total-energy drift, block standard errors, the number of independent density samples, and the drift of each energy term. Chain relaxation covers the dispersion of the chain radius of gyration, the chain center-of-mass displacement relative to $\langle R_g^2\rangle$, and the stationarity of the backbone torsion distribution. Structure covers nematic order, density homogeneity, and finite size. Chain relaxation and per-term energy drift bind  on the melt. A failing check that more sampling can fix triggers a bounded continuation of the same stage, and a cell that still fails a binding check is excluded from every property computed from it (SI Sections~S3 and S5).

\subsection{Property Calculation}

Density is averaged over the equilibrated plateau at the target temperature. $T_g$ is extracted from a stepwise cooling staircase that starts from the gated melt cell, by a smoothed-bilinear fit of density against temperature, and a fit that does not resolve a single transition is not reported. The bulk modulus is obtained from a Murnaghan equation-of-state fit over every simulated point of an NPT pressure ladder at the target temperature, and the volume-fluctuation modulus is kept as a cross-check. Chain dimensions, orientational order, and density homogeneity are computed on the gated cell (SI Section~S3).

\subsection{Benchmark Systems}

Seven amorphous homopolymers span nonpolar, aromatic, ester, ether, sulfone, and halide chemistries (Table~\ref{tab:benchmarks}). Each system was run as three replicates that share a frozen protocol and differ in their cell-construction and velocity seeds. The plan parameters of the replicates were compared key by key, and the seeds that reached EMC and LAMMPS match the plans and are distinct within each system (SI Section~S2).

Replicate~1 of each system was planned through the full path, including literature-critic review and adjudication, and replicates~2 and~3 reuse its frozen protocol. Acceptance criteria, comparable to tolerances used in MD polymer benchmarking,\cite{afzal2021highthroughput, Harish2024, maicas2021epoxyMD} are $|T_g^{\text{sim}} - T_g^{\text{exp}}| \leq 50$~K, a density error of at most 5\%, and a bulk-modulus error of at most 30\%. Where several measurements exist, a prediction inside their range has zero error and one outside is measured from the nearer end.

\begin{table}[ht]
\centering
\caption{Benchmark polymer systems and experimental reference properties at 300~K. A single value is given where one admissible measurement exists, and otherwise the range spans the admissible measurements, restricted to values whose measurement method and conditions are reported. $K_T$ is the isothermal bulk modulus, and --- marks a property without an experimental comparator, which is not graded.}
\label{tab:benchmarks}
\footnotesize
\begin{tabular}{@{}>{\raggedright\arraybackslash}p{2cm}*{3}{>{\centering\arraybackslash}p{\dimexpr(\textwidth-2cm-6\tabcolsep)/3\relax}}@{}}
\toprule
\textbf{Polymer} & \textbf{$T_g$ (K)} & \textbf{$\rho$ (g/cm$^3$)} & \textbf{$K_T$ (GPa)} \\
\midrule
PE    & 195--237\textsuperscript{a}\cite{boyer1973pe,gaur1981pe} & 0.855\cite{brandrup1999polymerhandbook,mark1999pdh} & 1.78--1.83\textsuperscript{b}\cite{hellwege1962compress,olabisi1975pvt} \\
PEG   & 213\cite{tormala1974pegtg} & 1.118--1.121\textsuperscript{c}\cite{mark2007pphandbook} & 2.26--2.42\textsuperscript{d}\cite{tsujita1973peg,becht1967peo} \\
PLLA  & 326--337\cite{mark1999pdh,garlotta2001,jamshidi1988pla} & 1.248\cite{garlotta2001,mikos1994plla} & --- \\
aPS   & 368--382\cite{brandrup1999polymerhandbook,wall1974ps} & 1.040--1.065\cite{brandrup1999polymerhandbook,mark1999pdh,mark2007pphandbook} & 3.59\cite{mark2007pphandbook} \\
sPVC  & 378--380\textsuperscript{e}\cite{mark1999pdh,ceccorulli1977pvc} & 1.385\textsuperscript{f}\cite{mark1999pdh,pham1974pvc} & 3.86\textsuperscript{g}\cite{hellwege1962compress} \\
PEEK  & 410--425\cite{blundell1983peek,mark1999pdh} & 1.263--1.265\cite{mark1999pdh} &  \\
PSU   & 458--459\cite{mark1999pdh,zoller1978psu} & 1.234--1.235\cite{zoller1978psu,mark2007pphandbook} & 4.64\cite{mark2007pphandbook} \\
\bottomrule
\multicolumn{4}{l}{\scriptsize \textsuperscript{a} The glass transition of PE is disputed. The range spans the two principal calorimetric assignments.}\\
\multicolumn{4}{l}{\scriptsize \textsuperscript{b} Room-temperature compressibility of semicrystalline PE extrapolated to the amorphous density.}\\
\multicolumn{4}{l}{\scriptsize \textsuperscript{c} Melt equation of state evaluated at 300~K, outside its measured range. No direct measurement at 300~K exists.}\\
\multicolumn{4}{l}{\scriptsize \textsuperscript{d} Liquid PEG 600 measured at 30.8~$^\circ$C, and linear PEO melt data extrapolated to 300~K.}\\
\multicolumn{4}{l}{\scriptsize \textsuperscript{e} No measurement exists for fully syndiotactic PVC.}\\
\multicolumn{4}{l}{\scriptsize \textsuperscript{f} Amorphous-phase density of PVC from the density--crystallinity relation.}\\
\multicolumn{4}{l}{\scriptsize \textsuperscript{g} Commercial glassy PVC. No tacticity-resolved measurement exists.}\\
\end{tabular}
\end{table}

\section{Results and Discussion}

\subsection{Density}
\label{sec:density}

\begin{figure}[H]
\centering
\includegraphics[width=0.7\textwidth]{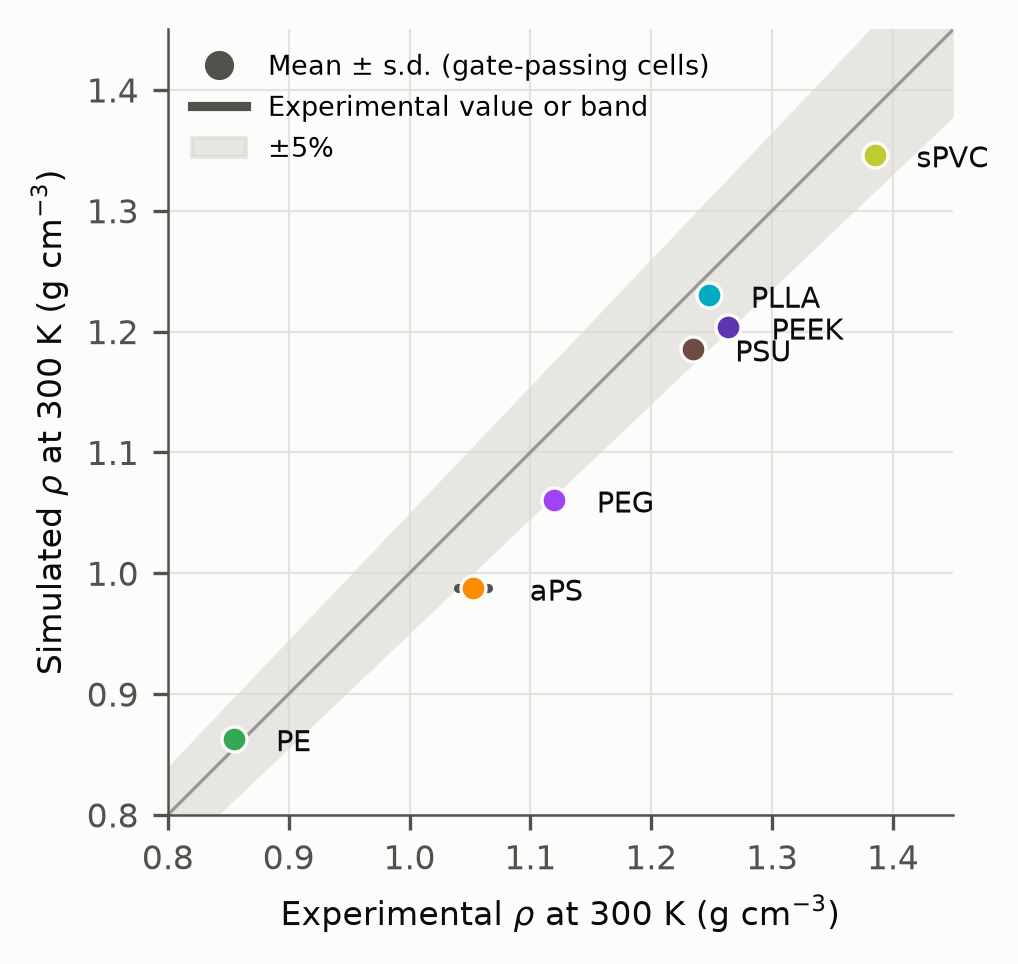}
\caption{Simulated versus experimental density at 300~K. Filled markers are means $\pm$ s.d. over gate-passing cells. Horizontal bars span the experimental range where several admissible measurements exist. The shaded region marks $\pm$5\% about parity.}
\label{fig:density_parity}
\end{figure}

Five of the seven systems meet the 5\% criterion, and aPS ($-$5.0\%) and PEG ($-$5.2\%) fall just outside it (Figure~\ref{fig:density_parity}). Every PCFF cell is less dense than its reference, by 1.4--5.2\%, while the united-atom PE cell is 0.9\% denser. For the glassy PCFF systems the attribution of this deficit to the force field remains a hypothesis. In a comparison study of PEG using different force fields, COMPASS brought the density within 0.7\% of the reference (SI Section~S4).

\subsection{Glass Transition Temperature}
\label{sec:tg_results}

\begin{figure*}[h]
\centering
\includegraphics[width=\textwidth]{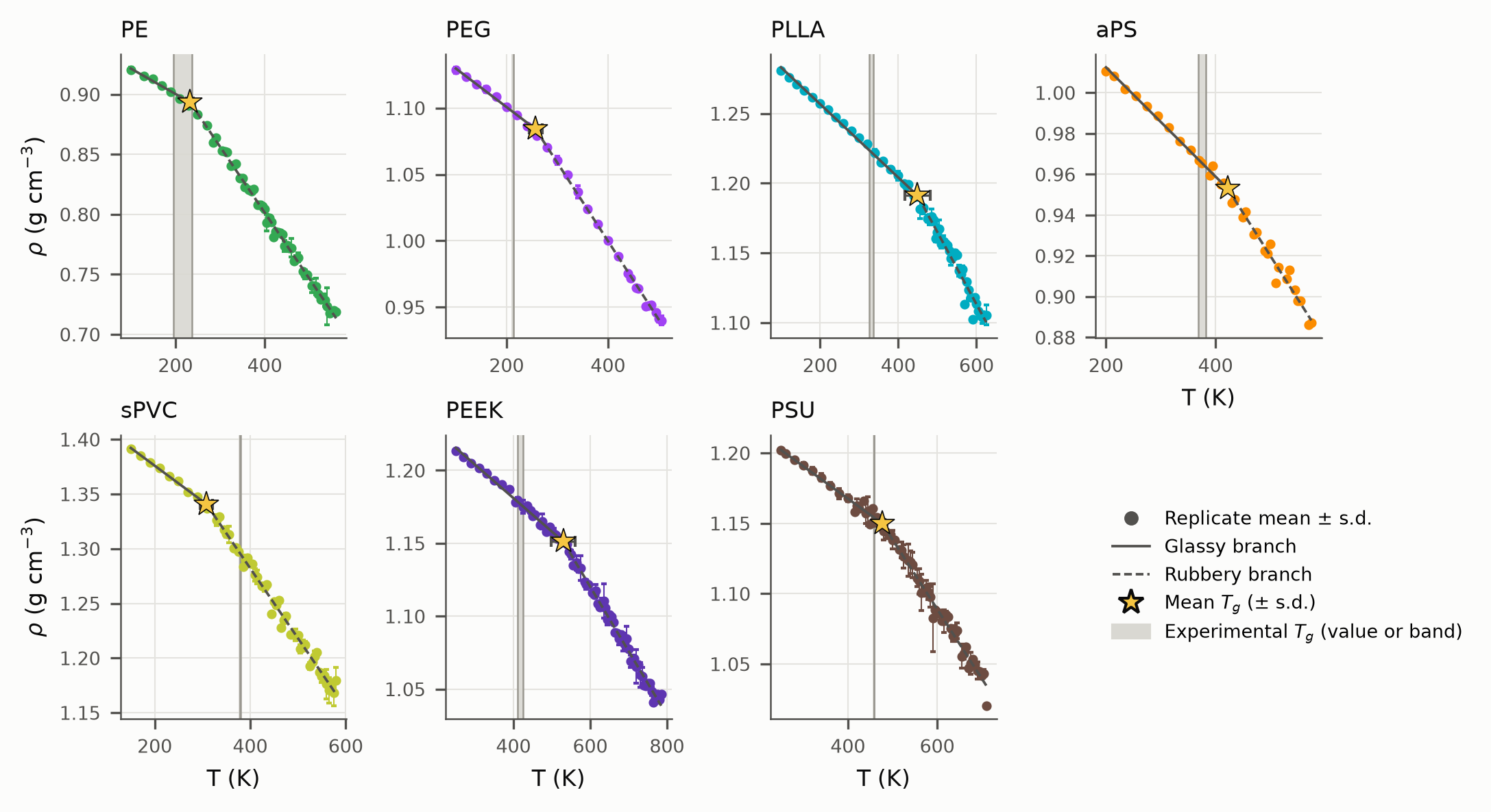}
\caption{Density versus temperature from the stepwise cooling sweep for each replicate. Solid and dashed lines are the glassy and rubbery branch fits, and dotted vertical lines mark reportable $T_g$ values. Gray bands give the experimental $T_g$ value or range.}
\label{fig:tg_curves}
\end{figure*}

Four of the seven systems meet the 50~K criterion (Figure~\ref{fig:tg_curves}). PLLA ($+$112.3~K) and PEEK ($+$104.8~K) are overestimated, and sPVC ($-$69.8~K) is underestimated against its syndiotactic-enriched reference. Six systems run above experiment, as expected from cooling rates many orders of magnitude faster than calorimetry. Four sweeps did not resolve a single transition and are excluded from the means. The mean absolute deviation is 55.6~K without calibration, whereas Afzal et al.\ report 27.5~K only after a linear calibration of predictions that were 79~K too high on average.\cite{afzal2021highthroughput} In a $T_g$ sensitivity test, cooling rate and temperature step shift $T_g$ by up to 39.6~K for PE and 101.5~K for PLLA (SI Section~S4).

\subsection{Bulk Modulus}

\begin{figure}[H]
\centering
\includegraphics[width=\columnwidth]{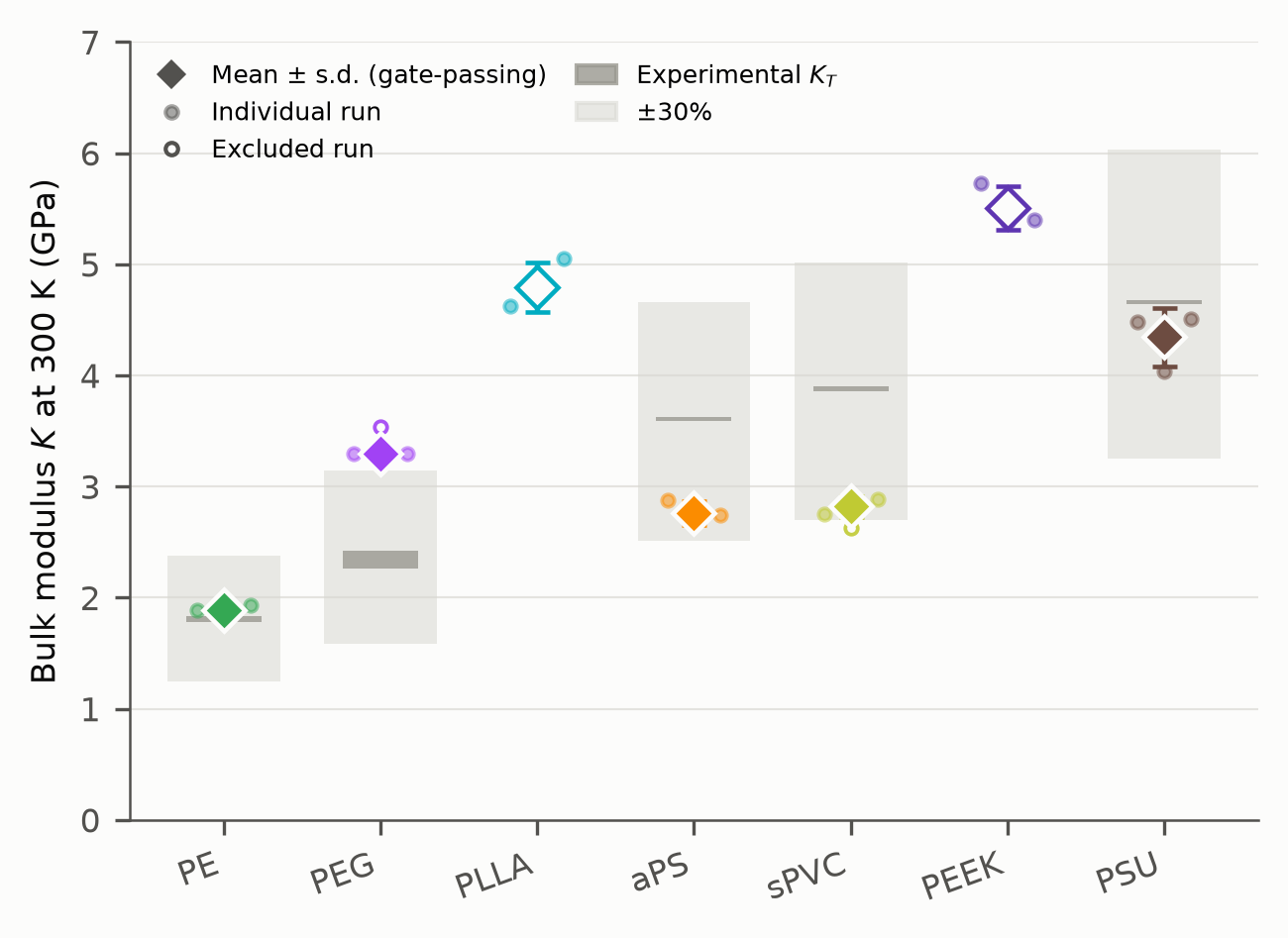}
\caption{Bulk modulus at 300~K from Murnaghan fits. Diamonds are means $\pm$ s.d. over gate-passing cells, and dots are individual runs, hollow when excluded. Dark gray marks the experimental isothermal $K_T$ value or range and light gray its $\pm$30\% tolerance. PLLA and PEEK have no isothermal experimental comparator and are not graded.}
\label{fig:bulk_modulus}
\end{figure}

Four of the five graded systems meet the 30\% criterion, and PEG is too stiff by 36.2\% (Figure~\ref{fig:bulk_modulus}). Replacing PCFF with COMPASS for PEG lowers its modulus by 21.8\%, to 14\% above the reference. The combined uncertainty from replicate scatter, pressure-ladder span, and equation-of-state form is 5--9\% for the glassy PCFF systems and PEG, 11\% for sPVC, and 23\% for PE. Controlled calculations on one sPVC cell changed $K$ by at most 10.1\% across barostat damping, production length, and ladder composition (SI Section~S4).

\subsection{Structural Validation and Energy Convergence}
\label{sec:structural}

Gate-passing cells show negligible orientational order, density heterogeneity below the random-packing floor, and near-Gaussian chain statistics, and none has a statistically significant energy drift (Table~\ref{tab:structdiag}).

\begin{table*}[ht]
\centering
\caption{Equilibration and conformational diagnostics for the seven benchmark systems at 300~K, as replicate-ensemble means $\pm$ 1 s.d. Gate criteria, and which of them bind at 300~K, are listed in SI Section~S3. Means are taken over every cell that passes the convergence gate, and the number of such cells is given in parentheses. Per-run values are in SI Section~S5.}
\label{tab:structdiag}
\small
\begin{tabular}{@{}lcccc>{\centering\arraybackslash}p{0.12\linewidth}c@{}}
\toprule
\textbf{Polymer} & \textbf{$\langle R_g\rangle$ (\AA{})} & \textbf{$R_g$ CV (\%)} & \textbf{MSID slope} & \textbf{$P_2$} & \textbf{$\rho$ CV (\%) (floor)} & \textbf{$E$ drift (\%)} \\
\midrule
PE (3)   & 33.2 $\pm$ 2.0 & 19 & 1.11 $\pm$ 0.04 & 0.035 $\pm$ 0.025 & 10 (19) & 1.94 $\pm$ 0.50 \\
PEG (2)  & 23.7 $\pm$ 0.4 & 19 & 1.02 $\pm$ 0.03 & 0.053 $\pm$ 0.002 & 20 (29) & 0.67 $\pm$ 0.11 \\
PLLA (3) & 16.2 $\pm$ 0.9 & 22 & 1.09 $\pm$ 0.04 & 0.013 $\pm$ 0.005 & 18 (24) & 0.04 $\pm$ 0.01 \\
aPS (3)  & 14.5 $\pm$ 0.4 & 13 & 1.30 $\pm$ 0.03 & 0.031 $\pm$ 0.013 & 24 (27) & 0.14 $\pm$ 0.09 \\
sPVC (2) & 21.7 $\pm$ 1.1 & 22 & 1.33 $\pm$ 0.01 & 0.030 $\pm$ 0.004 & 19 (32) & 2.27 $\pm$ 0.70 \\
PEEK (3) & 25.9 $\pm$ 0.6 & 22 & 1.26 $\pm$ 0.04 & 0.026 $\pm$ 0.014 & 16 (22) & 0.08 $\pm$ 0.05 \\
PSU (3)  & 19.9 $\pm$ 2.1 & 19 & 1.13 $\pm$ 0.06 & 0.010 $\pm$ 0.001 & 18 (23) & 0.11 $\pm$ 0.06 \\
\bottomrule
\end{tabular}

\end{table*}

\subsection{Agent Ablation on an Unseen Chemistry}
\label{sec:recovery}

\begin{figure}
\centering
\includegraphics[width=0.85\columnwidth]{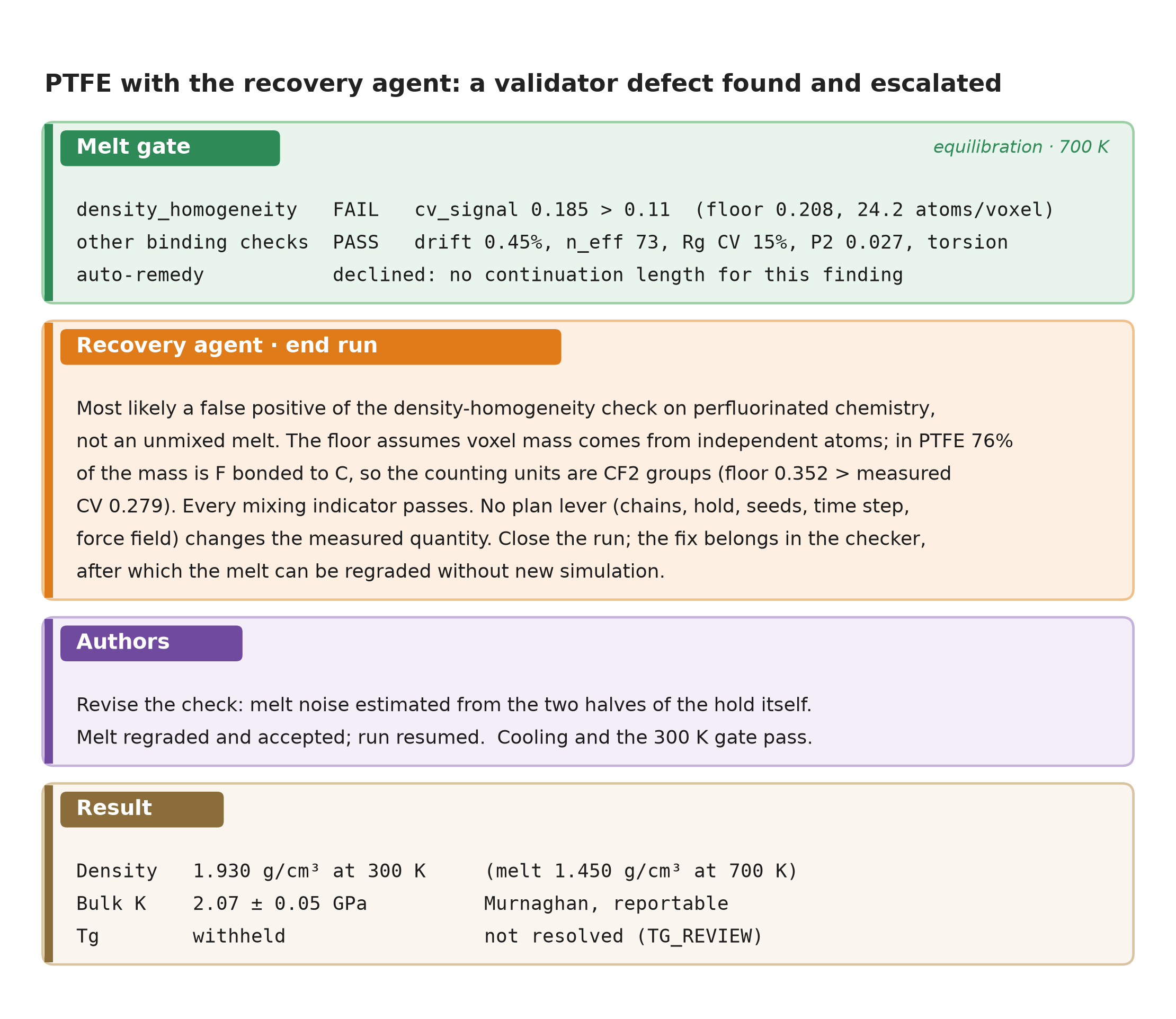}
\caption{a PolyJarvis run condensed from the records of the PTFE run with the recovery agent.}
\label{fig:conversation}
\end{figure}

PTFE, a system outside the validation set, was run by PolyJarvis with and without the recovery agent from the same plan and seeds, and by RadonPy's stock workflow as an external reference (Table~\ref{tab:recovery}, SI Section~S6). Both PolyJarvis melts failed only the density-homogeneity check, and without the agent the run stopped without a diagnosis. The agent attributed the failure to the check, whose counting-noise floor assumes independent atoms while most of PTFE's mass moves as bonded CF$_2$ groups. Finding no plan revision that changes the measured quantity, it ended the run and recommended correcting the check. The diagnosis was correct and specific to such chemistries, since all 21 campaign melts pass the original check and both PTFE melts pass the corrected one. After the authors revised the check, the agent run completed, with a second agent decision withholding an unresolved $T_g$. RadonPy's stock criteria contain no spatial-homogeneity test. Its equilibration ran to completion, but its own check reported the van der Waals energy as unconverged, so it recorded the cell as not equilibrated and skipped $T_g$ by its own rule. From the same trajectory it reports a density of 1.904~g\,cm$^{-3}$ and an isothermal bulk modulus of 2.65~GPa, against 1.930~g\,cm$^{-3}$ and 2.07~GPa for the agent arm, under a different force field and protocol, at a cost of about 28~h on 10 CPU cores. The ablation therefore shows correct diagnosis and escalation of a validator defect, not autonomous end-to-end recovery.

\begin{table}[ht]
\centering
\caption{Agent ablation on PTFE. Signals are those of the melt density-homogeneity check, whose limit is 0.11.}
\label{tab:recovery}
\footnotesize
\begin{tabular}{@{}>{\raggedright\arraybackslash}p{2.2cm} >{\raggedright\arraybackslash}p{2.4cm} >{\raggedright\arraybackslash}p{3.6cm}@{}}
\toprule
\textbf{Arm} & \textbf{Original $\to$ revised check} & \textbf{Outcome} \\
\midrule
PolyJarvis, no agent & 0.193 $\to$ 0.000 & stopped at escalation, 4.7~GPU-h \\
PolyJarvis, agent & 0.185 $\to$ 0.027 & check defect diagnosed, completed after correction, 14.3~GPU-h \\
RadonPy, stock & not measured & its own check failed, $T_g$ skipped, 28~h on 10 CPU cores \\
\bottomrule
\end{tabular}
\end{table}

\subsection{Validation Summary and Limitations}

\begin{table*}[ht]
\centering

\caption{Validation summary across the seven benchmark systems (replicate means over gate-passing cells). $\checkmark$ meets and $\times$ fails the criterion against an experimental reference, and --- marks a reference that is not graded. $T_g$ entries give $\Delta T_g$ in K, and density and bulk modulus entries give the relative error in percent.}
\label{tab:validation_summary}
\footnotesize
\setlength{\tabcolsep}{4pt}
\begin{tabular}{>{\raggedright\arraybackslash}p{0.1\linewidth}ccccccc}
\toprule
\textbf{Property (gate)} & \textbf{PE} & \textbf{PEG} & \textbf{PLLA} & \textbf{aPS} & \textbf{sPVC} & \textbf{PEEK} & \textbf{PSU} \\
\midrule
$\rho$ ($\pm$5\%)  & \checkmark\,$+$0.9 & $\times$\,$-$5.2 & \checkmark\,$-$1.4 & $\times$\,$-$5.0 & \checkmark\,$-$2.8 & \checkmark\,$-$4.7 & \checkmark\,$-$3.9 \\
$T_g$ ($\pm$50~K) & \checkmark\,0.0 & \checkmark\,$+$44.1 & $\times$\,$+$112.3 & \checkmark\,$+$40.2 & $\times$\,$-$69.8 & $\times$\,$+$104.8 & \checkmark\,$+$18.2 \\
$K$ ($\pm$30\%)   & \checkmark\,$+$3.0 & $\times$\,$+$36.2 & --- & \checkmark\,$-$23.0 & \checkmark\,$-$27.0 & --- & \checkmark\,$-$6.4 \\
\bottomrule
\end{tabular}
\end{table*}

Against experimental references, 13 of the 19 graded comparisons meet the acceptance criteria (Table~\ref{tab:validation_summary}). The failures are the under-density of aPS and PEG, the overestimated glass transitions of PLLA and PEEK, the underestimated glass transition of sPVC, and the overstiff PEG bulk modulus. The 21 replicates used 152 GPU-stage hours on an NVIDIA A800 workstation and 230 on a Quadro RTX 6000 workstation, including failed and superseded attempts (SI Section~S4).

\section{Conclusions}

We have presented PolyJarvis, a platform in which agents plan and make bounded recovery decisions while code owns simulation inputs, validation, and recovery limits, running end-to-end polymer MD workflows through EMC and LAMMPS. Across seven amorphous homopolymers, each run as three replicates sharing a frozen protocol with independent seeds, 13 of 19 graded comparisons meet the acceptance criteria. Density errors stay within 5.2\% in every system, while the glass transition is overestimated for the stiffest backbones. This accuracy suits screening and ranking, not the replacement of expert-curated protocols.

Because this work is a proof of concept, its scope is limited to amorphous cells, and broader validation on semicrystalline polymers, copolymers, and ionomer or polyelectrolyte systems remains future work, together with wider force-field coverage and a recovery benchmark scored on autonomous completion over repeated natural failures across systems.

\section*{Technology Use Disclosure}
The PolyJarvis system uses Anthropic's Claude large language model as the agent backbone.
Claude was also used to assist with grammar and typographical corrections during manuscript preparation.
All simulation results, analysis, and scientific conclusions have been carefully verified by the authors.

\section*{Data and Software Availability}

The code, run plans, prompts, MCP tool schemas, input scripts, seeds, final and intermediate structures, simulation logs, processed data, and the scripts that generate every table and figure are available at \url{https://github.com/arz-2/PolyJarvis} (tag acsami-r2, commit 5abfbea).

\begin{acknowledgement}

The authors thank the RadonPy development team (Y.\ Hayashi, R.\ Yoshida, and collaborators) for making RadonPy available as open-source software.
\end{acknowledgement}

\section*{Author Declarations}

\subsection*{Conflict of Interest}
The authors have no conflicts to disclose.

\subsection*{Author Contributions}
\textbf{Alexander Zhao:} Conceptualization, methodology, software development, simulation execution, data analysis, writing---original draft.
\textbf{Achuth Chandrasekhar:} Methodology, software development, writing---review and editing.
\textbf{Amir Barati Farimani:} Conceptualization, supervision, funding acquisition, writing---review and editing.

\begin{suppinfo}
Section~S1 describes the control plane, stage scripts, and MCP tool layer.
Section~S2 describes planning decisions and the simulation protocol, including classification, force-field selection and class priors, stereochemistry, system size, and the executed stage sequence.
Section~S3 gives the property calculation methods and the equilibration convergence gate.
Section~S4 reports the $T_g$ and bulk-modulus sensitivity calculations and computational cost.
Section~S5 gives per-run density, bulk modulus, $T_g$, and structural diagnostics.
Section~S6 describes the agent ablation on PTFE.
\end{suppinfo}

\bibliography{reference.bib}

\end{document}


\tableofcontents
\clearpage

\section{Control Plane and MCP Servers}
\label{sec:si_tools}

PolyJarvis separates scientific planning from execution. A planning step turns a repeat-unit SMILES and a set of target properties into a run plan that records every protocol decision with its admissible choices, the choice made, and its evidence. The plan is validated structurally and against policy before any simulation starts, and is then executed by deterministic stage scripts in a fixed order: cell construction, equilibration ending at a gated melt hold, cooling to the assessment temperature, the thermal and mechanical stages, and a summary.

The LLM (Claude\cite{anthropic2024claude}) enters at two points. In planning, a literature critic reviews the drafted plan against published MD studies of the same polymer and returns a verdict on the force-field decision with any suggested override. Then, an adjudicator accepts or declines each suggestion, and an accepted override is validated like any other plan edit. In execution, a recovery agent is consulted only when validation or a stage script returns a structured issue. Deterministic remedies for known issues are applied first, and the agent then chooses among bounded actions whose parameter revisions must pass the same bounds as the plan, within a fixed number of decisions per run and per stage.

Every simulation and analysis call is issued through one of three MCP\cite{anthropic2024mcp} servers, which are the only route from the stage scripts to a backend (Table~\ref{tab:si_servers}). Beyond ordinary API wrappers, MCP provides typed tool schemas, persisted multi-turn
context, asynchronous job lifecycle management, and structured error propagation.

\begin{table}[H]
\centering
\caption{The three MCP servers and the backend each one wraps.}
\label{tab:si_servers}
\small
\begin{tabular}{lll}
\toprule
\textbf{Server} & \textbf{Backend} & \textbf{Role} \\
\midrule
EMC builder & EMC\cite{intveld2003emc} & amorphous-cell construction and force-field typing \\
Molecule builder & RadonPy\cite{hayashi2022radonpy} & polymer classification and molecule utilities \\
LAMMPS engine & LAMMPS\cite{thompson2022lammps} & input generation, job chains, and analysis \\
\bottomrule
\end{tabular}
\end{table}

\section{Planning Decisions and Simulation Protocol}
\label{sec:si_decision_protocol}

\subsection{Polymer Classification}

The repeat unit is assigned one of 21 backbone classes derived from the PoLyInfo scheme\cite{otsuka2011polyinfo} by RadonPy's classifier, called unchanged. The class supplies a force-field prior and per-class protocol constants. Because the classifier matches on the extracted main chain, a separate functional-group profile of the whole repeat unit is compared with the class constants and any disagreement is reported to the critic as an advisory.

\subsection{Force-Field Selection}
\label{sec:si_ff}

Force-field assignment is fixed by polymer class: nonpolar hydrocarbon and diene backbones use TraPPE-UA,\cite{martin1998trappe} parameterized against phase-equilibrium data for these chemistries; polar, aromatic, and heteroatom backbones use PCFF,\cite{sun1994pcff} whose cross terms and bond-increment charges capture backbone stiffness and electrostatics; halogenated and siloxane backbones route to OPLS-AA.\cite{jorgensen1996opls}
A recent review confirms Class~II potentials are the most reliable choice for amorphous thermomechanical properties.\cite{nkepsumbitou2025}
Partial charges follow from the assigned force field---PCFF bond-increment charges, OPLS-AA library charges, or zero charges on the united-atom sites.

\begin{table}[H]
\centering
\caption{Trial EMC builds of each benchmark repeat unit and of PTFE under four force fields. The class prior is in bold. ``No typing rule'': EMC has no atom-typing rule for a group; ``missing parameters'': typing succeeds but a bonded or increment row is absent; ``element not typed'': the field defines no type for an element. A build shows only that the field can type the repeat unit, not that it is accurate for it.}
\label{tab:si_ff_coverage}
\small
\begin{tabular}{l l l l l}
\toprule
\textbf{System} & \textbf{PCFF} & \textbf{OPLS-AA} & \textbf{TraPPE-UA} & \textbf{COMPASS} \\
\midrule
PE & builds & builds & \textbf{builds} & builds \\
PEG & \textbf{builds} & missing parameters & builds & builds \\
PLLA & \textbf{builds} & missing parameters & no typing rule & missing parameters \\
aPS & \textbf{builds} & builds & builds & builds \\
sPVC & \textbf{builds} & missing parameters & no typing rule & element not typed \\
PEEK & \textbf{builds} & no typing rule & missing parameters & no typing rule \\
PSU & \textbf{builds} & no typing rule & no typing rule & no typing rule \\
PTFE & builds & \textbf{builds} & no typing rule & element not typed \\
\bottomrule
\end{tabular}
\end{table}

\subsection{Stereochemistry}
\label{sec:si_tacticity}

Stereochemistry is expressed only in the repeat-unit SMILES and is the same for every replicate of a system. aPS carries no stereo marker, so EMC places each backbone centre without stereocontrol (atactic). PLLA carries one defined centre whose configuration was checked by capping the repeat unit and comparing with L-lactic acid. sPVC is built from a two-unit repeat with opposite configurations at its two centres. 

\subsection{System Size}

Every cell contains 10 chains whose degree of polymerization gives about 5~kg\,mol$^{-1}$ per chain, for 4{,}940--8{,}000 atoms, or 3{,}600 united-atom sites for PE. Chain length is fixed by mass rather than by degree of polymerization because the chain-end depression of $T_g$ scales with molar mass,\cite{fox1950tg} and RadonPy likewise held molar mass nearly constant at 10 chains per cell, finding no significant box-size effect on density or other bulk properties.\cite{hayashi2022radonpy} For PEO, Wang et al.\ found $T_g$ independent of chain number outside a finite-size region, with scatter rising sharply below 11{,}240~g\,mol$^{-1}$ of total system mass and $T_g$ levelling off above 2{,}248~g\,mol$^{-1}$ per chain, thresholds these cells exceed about fourfold and twofold.\cite{wang2021mwtg} Because those thresholds were measured on PEO, any residual chain-length bias on stiffer backbones is not corrected but reported as an uncertainty.

\subsection{Executed Simulation Protocol}
\label{sec:si_protocol}

\begin{table}[H]
\centering
\caption{Executed protocol per system. $\rho_0$ is the packing density of the built cell; anneal and melt-hold durations are in ns; cooling is given as the number of 25~K blocks and the resulting rate; the $T_g$ staircase as its temperature range, 20~K step, and time per step; the Murnaghan series as pressure points $\times$ time per point.}
\label{tab:si_protocol}
\footnotesize
\begin{adjustbox}{max width=\textwidth}
\begin{tabular}{l l r r r r r l r l l}
\toprule
\textbf{System} & \textbf{FF} & $\rho_0$ & $T_{\mathrm{high}}$ (K), heat/hold & $T_{\mathrm{melt}}$ (K) & NVT hold & NPT hold & \textbf{Cooling} & 300~K hold & \textbf{$T_g$ staircase} & \textbf{Murnaghan} \\
\midrule
PE   & TraPPE-UA & 0.65 & 650, 1/1     & 550 & 5 & 2 & 10 blocks, 40~K/ns  & 5   & 550$\to$100~K, 0.5~ns & 6 $\times$ 1~ns \\
PEG  & PCFF      & 0.61 & 625, 1/1     & 500 & 5 & 2 & 8 blocks, 100~K/ns  & 2   & 500$\to$100~K, 0.2~ns & 5 $\times$ 0.5~ns \\
PLLA & PCFF      & 0.63 & 720, 1/1     & 620 & 5 & 2 & 13 blocks, 100~K/ns & 0.5 & 620$\to$100~K, 0.2~ns & 5 $\times$ 0.5~ns \\
aPS  & PCFF      & 0.53 & 650, 1/1     & 573 & 5 & 6 & 11 blocks, 100~K/ns & 0.5 & 573$\to$200~K, 0.2~ns & 5 $\times$ 0.5~ns \\
sPVC & PCFF      & 0.55 & 630, 0.5/0.5 & 571 & 5 & 4 & 11 blocks, 100~K/ns & 0.5 & 571$\to$150~K, 0.2~ns & 5 $\times$ 0.5~ns \\
PEEK & PCFF      & 0.66 & 870, 1/1     & 770 & 5 & 2 & 19 blocks, 100~K/ns & 0.5 & 770$\to$250~K, 0.2~ns & 5 $\times$ 0.5~ns \\
PSU  & PCFF      & 0.62 & 800, 1/1     & 700 & 5 & 2 & 16 blocks, 100~K/ns & 0.5 & 700$\to$250~K, 0.2~ns & 5 $\times$ 0.5~ns \\
\bottomrule
\end{tabular}
\end{adjustbox}
\end{table}

\section{Property Calculation Methods}
\label{sec:si_property_methods}

\subsection{Density}

Density is the plateau-averaged mass-to-volume ratio of the NPT hold at 300~K and 1~atm that ends cooling. Rather than a fixed final-half window, the reported value is the mean over the longest stable tail, located by a reverse cumulative-mean scan that extends backward until adding a row shifts the running mean by more than one standard error. Convergence is quantified by the autocorrelation time, the effective sample size, and a block standard error.\cite{chodera2016equilibration} Where a stage was extended by restart continuation, the continued segments are graded as one continuous run.

\subsection{Glass Transition Temperature ($T_g$)}

$T_g$ is determined from a single NPT staircase per run that starts from the gated melt cell at $T_{\mathrm{melt}}$ and descends in 20~K steps at 1~atm (Table~\ref{tab:si_protocol}). Each step lasts 0.2~ns for the PCFF systems (100~K/ns) and 0.5~ns for PE (40~K/ns). The density of each step is averaged over the stable tail of its plateau, and a step whose density window is unstable or has too few independent samples receives an additional hold at that temperature.

$T_g$ is extracted from a smoothed-bilinear (hyperbola) fit of the density--temperature data\cite{patrone2016uncertainty}:
\begin{equation}
\rho(T) = \rho_0 + \bar{m}\,(T - T_g) + \delta\sqrt{(T - T_g)^2 + c^2},
\end{equation}
whose asymptotic slopes give the glassy and rubbery branches and whose vertex gives $T_g$; $c$ is the crossover half-width. Fit quality follows $R^2$: EXCELLENT ($\ge 0.995$), GOOD ($\ge 0.98$), ACCEPTABLE ($\ge 0.95$), POOR otherwise. A $T_g$ is reportable unless the fit is POOR or invalid, or the transition is not resolved. For a segmented-bilinear fit this is breakpoint ambiguity: competing splits within 10\% of the best residual spread by more than 20~K, or twice the fit uncertainty exceeding 20~K. Otherwise it is a method gap: the primary and alternative fits differ by more than 20~K. Non-reportable fits are excluded from $T_g$ means.

\subsection{Bulk Modulus}

The bulk modulus is obtained from the Murnaghan equation of state\cite{murnaghan1944}
\begin{equation}
P(V) = \frac{B_0}{B_0'}\left[\left(\frac{V_0}{V}\right)^{B_0'} - 1\right],
\end{equation}
fit to the mean volumes of an NPT pressure series at 300~K, each averaged over the last half of its production window, reporting $K \equiv B_0$, $B_0'$, $R^2$, and the fit standard error. The fit script flags a pressure point whose volume change relative to its neighbour exceeds three times the median interval, and selects the smallest trimmed window that converges with volume decreasing monotonically in pressure; the excluded points are recorded. The isothermal volume-fluctuation estimator $K = k_B T \langle V\rangle / \langle(\delta V)^2\rangle$\cite{allen2017computer} is reported as a cross-check. The reported moduli use every simulated pressure.

\subsection{Structural Characterization}
\label{sec:si_structural_methods}

Structural diagnostics are computed on the trajectory of each gated cell, with coordinates unwrapped and backbone atoms selected by LAMMPS atom type. The per-run values reported in Section~\ref{sec:si_results} are those of the cell at the assessment temperature.

\subsubsection{Radius of Gyration and End-to-End Distance}

The mass-weighted radius of gyration per chain is
\begin{equation}
R_g = \left[\frac{\sum_{i=1}^{N} m_i(\mathbf{r}_i - \mathbf{r}_{\mathrm{cm}})^2}{\sum_{i=1}^{N} m_i}\right]^{1/2},
\end{equation}
where the sum runs over all $N$ atoms of the chain and $\mathbf{r}_{\mathrm{cm}}$ is the chain centre of mass. The chain-mean $\langle R_g\rangle$ and the coefficient of variation across chains are reported. The end-to-end distance is $R = \|\mathbf{r}_{\mathrm{last}} - \mathbf{r}_{\mathrm{first}}\|$ per chain per frame, with termini ordered along the backbone.

\subsubsection{Mean Square Internal Distance}

The mean square internal distance $\langle R^2(n)\rangle$\cite{auhl2003equilibration} as a function of backbone bond separation $n$ is
\begin{equation}
\langle R^2(n)\rangle = \frac{1}{M_n}\sum_{i}\left|\mathbf{r}_{i+n} - \mathbf{r}_i\right|^2,
\end{equation}
where $M_n$ is the number of bond pairs separated by exactly $n$ bonds. The diagnostic reports the slope of $\log\langle R^2(n)\rangle$ against $\log n$.

\subsubsection{Nematic Order Parameter}

The nematic order parameter $P_2 = \frac{1}{2}\left\langle 3\cos^2\theta - 1\right\rangle$ is evaluated as the largest eigenvalue of the Saupe tensor $\mathbf{Q} = \frac{1}{2N}\sum_i\left(3\,\mathbf{u}_i\otimes\mathbf{u}_i - \mathbf{I}\right)$ over unit backbone bond vectors $\mathbf{u}_i$, averaged over frames.

\subsubsection{Density Homogeneity}

The cell is divided into voxels, and the coefficient of variation of voxel mass density is compared with its counting-noise floor for the same number of atoms per voxel; the gate reads the CV remaining after the floor is subtracted. A trajectory-based form of the noise floor, described in Section~\ref{sec:si_recovery}, is used where noted.

\subsubsection{Chain Diffusion}

Chain centre-of-mass mean-square displacement is fit to $\mathrm{MSD}(\tau) = A\,\tau^{\alpha}$; a cell is flagged as kinetically trapped when the maximum displacement falls below $\langle R_g^2\rangle$.

\subsection{Equilibration Convergence Gate}
\label{sec:si_gate}

The gate is evaluated on two cells per run with the same checker and classifier. The melt gate reads the last half of the melt-hold log, the fixed-volume melt trajectory for chain displacement, and the melt-hold trajectory for structure. The assessment gate reads the corresponding records of the final cell at the assessment temperature. Table~\ref{tab:si_gate} lists every check, its criterion, and whether it binds in each clause.

\begin{table}[H]
\centering
\caption{Convergence-gate checks. B = binding, A = advisory (reported, never blocks). Drift criteria require both the magnitude threshold and an autocorrelation-corrected $p < 0.01$ to fail.}
\label{tab:si_gate}
\scriptsize
\begin{adjustbox}{max width=\textwidth}
\begin{tabular}{l p{6.6cm} c c c}
\toprule
\textbf{Check} & \textbf{Passes when} & \textbf{Melt} & \textbf{Glassy} & \textbf{Rubbery} \\
\midrule
Density drift & drift $\le$ 1\% of mean & B & B & A \\
Total-energy drift & drift $\le$ 1\% of mean & B & B & B \\
Density, energy block SEM & block standard error $<$ 1\% (10 blocks) & B & B & B \\
Independent density samples & $n_{\mathrm{eff}} \ge 20$ & B & B & B \\
Per-term energy drift & each of bond, angle, dihedral, vdW, Coulomb, $k$-space: drift $\le$ 1 standard deviation of that term's own fluctuation & B & A & A \\
$R_g$ dispersion & coefficient of variation across chains $<$ 30\% & B & A & A \\
Chain displacement & centre-of-mass MSD $\ge \langle R_g^2\rangle$ and not kinetically trapped & B & --- & --- \\
Torsion stationarity & Jensen--Shannon divergence between consecutive blocks of the backbone dihedral distribution $<$ 0.05 & B & A & A \\
Nematic order & $P_2 < 0.10$ & B & B & A \\
Density homogeneity & voxel mass-density CV after subtracting counting noise $\le$ 0.11 & B & B & B \\
Finite size & $L \ge 2\,r_c$; chain self-imaging binds below $L/2R_g = 0.5$ & B & B & B \\
MSID slope, MSD trap flag, residual stress, $C(t)$ decay & --- & A & A & A \\
\bottomrule
\end{tabular}
\end{adjustbox}
\end{table}

A clause with no measured binding check returns FAIL. Otherwise the verdict is PASS when no binding check fails; EXTEND when every failing check is one that more sampling can remedy (drift, SEM, sample count, per-term energy drift, $R_g$ dispersion, chain displacement, torsion), which triggers a restart continuation of the same stage sized from the measured deficit and capped at two applications per check; STRUCTURAL\_FAIL when density homogeneity, nematic order, or finite size fails; and FAIL otherwise. For glassy cells the melt-to-glass contraction is additionally compared with the cell's own thermal-expansion prediction.

\section{Sensitivity Calculations and Computational Cost}
\label{sec:si_sensitivity}

\subsection{$T_g$ Sensitivity}

Each leg reran only the $T_g$ staircase from an anchor replicate's gated melt cell with one setting changed, and a shift counts as sensitive when it exceeds the pooled within-system replicate standard deviation, 19.5~K (10 degrees of freedom), a threshold fixed before the legs were run. For PE (PE\_1: 40~K\,ns$^{-1}$, 20~K step, $T_g$ = 234.2~K), cooling at 20 and 10~K\,ns$^{-1}$ shifted $T_g$ by $-$20.8 and $-$11.1~K, and at 20~K\,ns$^{-1}$ steps of 10 and 40~K shifted it by $+$39.6 and $-$5.0~K relative to the 20~K step at that rate. For PLLA (PLLA\_1: 100~K\,ns$^{-1}$, 20~K step, $T_g$ = 483.6~K), a 40~K step shifted $T_g$ by $-$101.5~K, and the 50~K\,ns$^{-1}$ leg gave no reportable $T_g$ (hyperbolic and bilinear estimates 78.5~K apart). Refitting every accepted staircase with alternative procedures, after confirming that each stored $T_g$ is reproduced exactly, moved one system mean beyond the threshold: averaging only the last 25\% of each step lowered $T_g$ for aPS by 26.4~K, in its single reportable run (Table~\ref{tab:si_tg_fit}).

\begin{table}[H]
\centering
\caption{$T_g$ fit-procedure sensitivity over the 21 campaign staircases (no new simulation). $|\Delta T_g|$ is the largest system-mean shift from the live fit over the 17 runs whose live fit is reportable; the threshold is 19.5~K.}
\label{tab:si_tg_fit}
\footnotesize
\begin{adjustbox}{max width=\textwidth}
\begin{tabular}{@{}l>{\centering\arraybackslash}p{1.3cm}>{\centering\arraybackslash}p{2.2cm}>{\centering\arraybackslash}p{1.9cm}c@{}}
\toprule
Procedure & $|\Delta T_g|$ (K) & $|\Delta T_g|$, reportable under both (K) & Fits made non-reportable & Sensitive \\
\midrule
Segmented bilinear only & 8.7 & 7.6 & 9/17 & no \\
Smoothed bilinear (hyperbola) only & 0.0 & 0.0 & 0/17 & no \\
Last 25\% of each step averaged & 26.4 & 26.4 & 1/17 & yes \\
Last 75\% of each step averaged & 17.0 & 17.0 & 3/17 & no \\
Two hottest temperature groups removed & 53.7 & 14.3 & 2/17 & no\textsuperscript{a} \\
Two coldest temperature steps removed & 43.6 & 12.2 & 3/17 & no\textsuperscript{a} \\
\bottomrule
\end{tabular}
\end{adjustbox}

\raggedright\scriptsize\textsuperscript{a}\,Exceeds 19.5~K only through fits the alternative leaves unresolved (TG\_REVIEW).
\end{table}

\subsection{Bulk-Modulus Robustness}

Every executed pressure series was refitted without new simulation, after reproducing each stored $B_0$ exactly from the logs. The span dispersion is the standard deviation of Murnaghan refits over every contiguous sub-ladder of at least four pressures and every leave-one-out set; the form dispersion is the standard deviation across Murnaghan, third-order Birch--Murnaghan, and Vinet fits to the same volumes. Both are combined in quadrature with the replicate standard deviation (Table~\ref{tab:si_bm_dispersion}). All moduli are reported on all simulated points, so that the fit script's automatic point exclusion cannot select a different ladder for different cells; that exclusion alone moves $K$ by up to 30\% (Table~\ref{tab:si_bm_dispersion}, footnote). For PE, Murnaghan, Vinet, and Birch--Murnaghan fits to identical points give 1.83--1.93, 1.55--1.65, and 1.09--1.22~GPa, all with $R^2 \geq 0.9997$.

\begin{table}[H]
\centering
\caption{Controlled bulk-modulus calculations on the sPVC\_3 cell. $B_0$ is the Murnaghan fit on all points of each leg; $\Delta$ is relative to the all-point baseline.}
\label{tab:si_bm_dispersion}
\footnotesize
\begin{tabular}{@{}l l l c c@{}}
\toprule
Leg & Setting & Pressures (atm) & $B_0$ (GPa) & $\Delta$ (\%) \\
\midrule
Baseline & damping 1000~fs, 500{,}000 steps & $-$1000, 0, 1500, 3000, 5000 & 2.888 & 0.0 \\
Barostat damping & 250~fs & $-$1000, 0, 1500, 3000, 5000 & 2.770 & $-$4.1 \\
Barostat damping & 4000~fs & $-$1000, 0, 1500, 3000, 5000 & 2.957 & $+$2.4 \\
Production & 250{,}000 steps & $-$1000, 0, 1500, 3000, 5000 & 3.057 & $+$5.9 \\
Production & 1{,}000{,}000 steps & $-$1000, 0, 1500, 3000, 5000 & 2.797 & $-$3.2 \\
Span & $\pm$1000~atm & $-$1000, $-$500, 0, 500, 1000 & 2.963 & $+$2.6 \\
Ladder density & 7 points & $-$1000, 0, 750, 1500, 3000, 4000, 5000 & 2.900 & $+$0.4 \\
Ceiling & 15{,}000~atm & $-$1000, 0, 3000, 7000, 15000 & 3.024 & $+$4.7 \\
No tension point & 1--5000~atm & 1, 1250, 2500, 3750, 5000 & 3.180 & $+$10.1 \\
Tension only & $-$1000 to 0~atm & $-$1000, $-$500, $-$250, 0 & 2.694 & $-$6.7 \\
\bottomrule
\end{tabular}
\end{table}

Doubling the chain count of a rubbery COMPASS PEG cell from 10 to 20 changed $B_0$ by $+$0.5\%.

\subsection{Force-Field Change for PEG}
\label{sec:si_peg_compass}

PEG was rerun with the force field changed from PCFF to COMPASS and everything else held fixed: repeat unit, cell size (DP 114, 10 chains, 8000 atoms), the frozen protocol, cutoff, and the cell-construction and velocity seeds of each PCFF replicate. The field-derived settings (bond-increment charges, PPPM, 1~fs timestep) are the same for both fields. The run passed the melt and 300~K gates.

\begin{table}[H]
\centering
\caption{PEG under PCFF and COMPASS (seed pair 2). References: amorphous PEO density 1.1183--1.1205~g\,cm$^{-3}$ and $K_T$ = 2.26--2.42~GPa.}
\label{tab:si_peg_compass}
\footnotesize
\begin{tabular}{lccc}
\toprule
 & PEG\_2 (PCFF) & COMPASS & COMPASS $-$ PCFF \\
\midrule
$\rho$, 300~K (g\,cm$^{-3}$) & 1.0600 & 1.1278 & $+$6.4\% \\
$\rho$, 500~K melt (g\,cm$^{-3}$) & 0.9378 & 0.9606 & $+$2.4\% \\
Murnaghan $K$ (GPa) & 3.53 $\pm$ 0.11 & 2.76 $\pm$ 0.07 & $-$21.8\% \\
$B_0'$ & 8.99 & 9.64 & \\
Fluctuation $K$ (GPa) & 3.63 & 2.60 & $-$28.4\% \\
\bottomrule
\end{tabular}
\end{table}

\subsection{Computational Cost}
\label{sec:si_cost}

Runs for aPS, sPVC, and PLLA were executed on a workstation with four NVIDIA Quadro RTX 6000 GPUs, and runs for PE, PEG, PEEK, and PSU on a workstation with four NVIDIA A800 GPUs, one GPU per simulation stage. Table~\ref{tab:si_walltimes} gives the engine time of the GPU stages (equilibration, cooling, thermal, mechanical) summed over every attempt, including failed and superseded attempts.

\begin{table}[H]
\centering
\caption{GPU-stage engine time per system (all attempts).}
\label{tab:si_walltimes}
\small
\begin{tabular}{l l l r r r}
\toprule
\textbf{System} & \textbf{FF} & \textbf{GPU} & \textbf{Per replicate (h)} & \textbf{Failed/superseded (h)} & \textbf{Total (h)} \\
\midrule
PE   & TraPPE-UA & A800     & 3.6--7.1   & 6.1  & 17.0 \\
PEG  & PCFF      & A800     & 12.5--17.4 & 4.9  & 43.3 \\
PEEK & PCFF      & A800     & 14.6--16.3 & 0.0  & 46.9 \\
PSU  & PCFF      & A800     & 14.2--15.6 & 0.0  & 45.3 \\
PLLA & PCFF      & RTX 6000 & 26.0--33.4 & 5.6  & 85.6 \\
aPS  & PCFF      & RTX 6000 & 30.8--31.7 & 0.0  & 93.6 \\
sPVC & PCFF      & RTX 6000 & 16.7--17.3 & 0.0  & 50.7 \\
\midrule
\multicolumn{5}{l}{Total, A800} & 152 \\
\multicolumn{5}{l}{Total, RTX 6000} & 230 \\
\bottomrule
\end{tabular}
\end{table}

\section{Results}
\label{sec:si_results}

Per-run values for all 21 runs are given below.

\subsection{Density and Bulk Modulus}

\scriptsize
\begin{longtable}{l r r r r r r r l l c}
\caption{Per-run density at 300~K and in the melt, Murnaghan bulk modulus, and the gate disposition of each cell under the frozen gate version: $K \pm$ fit standard error, $B_0'$, $R^2$, number of pressure points fitted, and the volume-fluctuation cross-check $K_{\mathrm{fluct}}$. The last three columns give the verdict of each gated cell under the frozen gate version and whether the run's 300~K properties are included. Two cells fail a binding thermodynamic check under the frozen regrade and are excluded.} \label{tab:si_rho_k} \\
\toprule
\textbf{Run} & \textbf{$\rho_{300}$} & \textbf{$\rho_{\mathrm{melt}}$} & \textbf{$K$ (GPa)} & \textbf{$B_0'$} & \textbf{$R^2$} & \textbf{Pts} & \textbf{$K_{\mathrm{fluct}}$} & \textbf{Melt} & \textbf{300~K} & \textbf{Incl.} \\
\midrule
\endfirsthead
\toprule
\textbf{Run} & \textbf{$\rho_{300}$} & \textbf{$\rho_{\mathrm{melt}}$} & \textbf{$K$ (GPa)} & \textbf{$B_0'$} & \textbf{$R^2$} & \textbf{Pts} & \textbf{$K_{\mathrm{fluct}}$} & \textbf{Melt} & \textbf{300~K} & \textbf{Incl.} \\
\midrule
\endhead
\bottomrule
\endfoot
PE\_1 & 0.8637 & 0.7189 & 1.890 $\pm$ 0.120 & 8.38 & 0.9999 & 6 & 1.77 & PASS & PASS & yes \\
PE\_2 & 0.8592 & 0.7168 & 1.830 $\pm$ 0.157 & 8.27 & 0.9998 & 6 & 1.47 & PASS & PASS & yes \\
PE\_3 & 0.8661 & 0.7189 & 1.933 $\pm$ 0.137 & 8.25 & 0.9998 & 6 & 1.69 & PASS & PASS & yes \\
PEG\_1 & 1.0579 & 0.9391 & 3.297 $\pm$ 0.120 & 9.42 & 0.9999 & 5 & 3.28 & PASS & PASS & yes \\
PEG\_2 & 1.0600 & 0.9378 & 3.534 $\pm$ 0.113 & 8.99 & 0.9999 & 5 & 3.63 & PASS & FAIL (E drift) & no \\
PEG\_3 & 1.0631 & 0.9375 & 3.296 $\pm$ 0.123 & 9.58 & 0.9999 & 5 & 3.82 & PASS & PASS & yes \\
PLLA\_1 & 1.2341 & 1.1022 & 4.621 $\pm$ 0.167 & 9.78 & 0.9996 & 5 & 5.84 & PASS & PASS & yes \\
PLLA\_2 & 1.2270 & 1.1018 & 4.717 $\pm$ 0.059 & 11.40 & 1.0000 & 5 & 4.60 & PASS & PASS & yes \\
PLLA\_3 & 1.2289 & 1.1034 & 5.048 $\pm$ 0.172 & 8.54 & 0.9997 & 5 & 5.15 & PASS & PASS & yes \\
aPS\_1 & 0.9854 & 0.8859 & 2.880 $\pm$ 0.181 & 7.90 & 0.9988 & 5 & 3.03 & PASS & PASS & yes \\
aPS\_2 & 0.9882 & 0.8856 & 2.669 $\pm$ 0.198 & 9.22 & 0.9984 & 5 & 2.50 & PASS & PASS & yes \\
aPS\_3 & 0.9891 & --- & 2.740 $\pm$ 0.094 & 9.70 & 0.9997 & 5 & 3.44 & PASS & PASS & yes \\
sPVC\_1 & 1.3452 & 1.1714 & 2.750 $\pm$ 0.094 & 10.01 & 0.9997 & 5 & 3.26 & PASS & PASS & yes \\
sPVC\_2 & 1.3413 & 1.1713 & 2.627 $\pm$ 0.198 & 10.24 & 0.9984 & 5 & 3.36 & PASS & FAIL (E SEM) & no \\
sPVC\_3 & 1.3472 & 1.1716 & 2.888 $\pm$ 0.245 & 9.81 & 0.9980 & 5 & 3.27 & PASS & PASS & yes \\
PEEK\_1 & 1.2055 & 1.0399 & 5.729 $\pm$ 0.232 & 7.38 & 0.9999 & 5 & 5.06 & PASS & PASS & yes \\
PEEK\_2 & 1.2017 & 1.0401 & 5.383 $\pm$ 0.059 & 8.28 & 1.0000 & 5 & 5.52 & PASS & PASS & yes \\
PEEK\_3 & 1.2038 & 1.0401 & 5.402 $\pm$ 0.118 & 8.09 & 1.0000 & 5 & 5.63 & PASS & PASS & yes \\
PSU\_1 & 1.1843 & 1.0409 & 4.483 $\pm$ 0.115 & 7.76 & 1.0000 & 5 & 3.72 & PASS & PASS & yes \\
PSU\_2 & 1.1877 & 1.0428 & 4.038 $\pm$ 0.121 & 8.86 & 0.9999 & 5 & 3.77 & PASS & PASS & yes \\
PSU\_3 & 1.1846 & 1.0415 & 4.511 $\pm$ 0.041 & 8.24 & 1.0000 & 5 & 4.71 & PASS & PASS & yes \\
\end{longtable}
\normalsize

\subsection{Glass Transition Temperature}

\scriptsize
\begin{longtable}{l r r r l r l}
\caption{Per-run $T_g$ from the smoothed-bilinear fit: uncertainty, $R^2$, fit quality, crossover half-width $c$, and reportability verdict.} \label{tab:si_tg} \\
\toprule
\textbf{Run} & \textbf{$T_g$ (K)} & \textbf{$\pm$ (K)} & \textbf{$R^2$} & \textbf{Fit} & \textbf{$c$ (K)} & \textbf{Verdict} \\
\midrule
\endfirsthead
\toprule
\textbf{Run} & \textbf{$T_g$ (K)} & \textbf{$\pm$ (K)} & \textbf{$R^2$} & \textbf{Fit} & \textbf{$c$ (K)} & \textbf{Verdict} \\
\midrule
\endhead
\bottomrule
\endfoot
PE\_1 & 234.2 & 13.3 & 0.9972 & Excellent & 22.8 & Reportable \\
PE\_2 & 239.8 & 10.0 & 0.9966 & Excellent & --- & Reportable \\
PE\_3 & 223.5 & 21.2 & 0.9966 & Excellent & 37.4 & Reportable \\
PEG\_1 & 259.6 & 9.9 & 0.9990 & Excellent & 24.4 & Reportable \\
PEG\_2 & 268.4 & 12.1 & 0.9991 & Excellent & 58.7 & Reportable \\
PEG\_3 & 243.8 & 9.3 & 0.9997 & Excellent & 67.0 & Reportable \\
PLLA\_1 & 483.6 & 21.5 & 0.9974 & Excellent & 82.9 & Reportable \\
PLLA\_2 & 421.0 & 26.3 & 0.9933 & Good & 84.2 & Reportable \\
PLLA\_3 & 443.2 & 19.9 & 0.9955 & Excellent & 84.5 & Reportable \\
aPS\_1 & 422.4 & 56.2 & 0.9938 & Good & 128.2 & Reportable \\
aPS\_2 & 348.6 & 16.5 & 0.9966 & Excellent & --- & Review (breakpoint ambiguity) \\
aPS\_3 & 346.7 & 9.4 & 0.9949 & Good & --- & Review (breakpoint ambiguity) \\
sPVC\_1 & 293.9 & 21.2 & 0.9931 & Good & 17.5 & Reportable \\
sPVC\_2 & 314.3 & 16.8 & 0.9937 & Good & --- & Reportable \\
sPVC\_3 & 316.4 & 41.6 & 0.9945 & Good & 98.0 & Reportable \\
PEEK\_1 & 551.9 & 9.0 & 0.9939 & Good & 4.8 & Reportable \\
PEEK\_2 & 572.4 & 19.6 & 0.9841 & Good & --- & Review (breakpoint ambiguity) \\
PEEK\_3 & 507.6 & 7.7 & 0.9967 & Excellent & 4.8 & Reportable \\
PSU\_1 & 481.3 & 14.7 & 0.9849 & Good & 25.6 & Reportable \\
PSU\_2 & 525.0 & 14.2 & 0.9851 & Good & 0.0 & Review (method gap) \\
PSU\_3 & 473.4 & 10.2 & 0.9943 & Good & --- & Reportable \\
\end{longtable}
\normalsize

\subsection{Structural Diagnostics}

\scriptsize
\begingroup\setlength{\tabcolsep}{3.8pt}
\begin{longtable}{l r r r r r r r r l}
\caption{Per-run structural and convergence diagnostics of the 300~K cell: mean radius of gyration and its coefficient of variation across chains, mean end-to-end distance, MSID slope, nematic order $P_2$, voxel density CV with its counting-noise floor in parentheses, total-energy drift of the graded window (continuations read as one run), and independent density samples. Seeds are the EMC packing seed / velocity seed.} \label{tab:si_struct} \\
\toprule
\textbf{Run} & \textbf{$\langle R_g\rangle$ (\AA)} & \textbf{$R_g$ CV (\%)} & \textbf{$\langle R_{ee}\rangle$ (\AA)} & \textbf{MSID} & \textbf{$P_2$} & \textbf{$\rho$ CV (\%)} & \textbf{$E$ drift (\%)} & \textbf{$n_{\mathrm{eff}}$} & \textbf{Seeds} \\
\midrule
\endfirsthead
\toprule
\textbf{Run} & \textbf{$\langle R_g\rangle$ (\AA)} & \textbf{$R_g$ CV (\%)} & \textbf{$\langle R_{ee}\rangle$ (\AA)} & \textbf{MSID} & \textbf{$P_2$} & \textbf{$\rho$ CV (\%)} & \textbf{$E$ drift (\%)} & \textbf{$n_{\mathrm{eff}}$} & \textbf{Seeds} \\
\midrule
\endhead
\bottomrule
\endfoot
PE\_1 & 31.3 & 22 & 73.8 & 1.07 & 0.021 & 10 (19) & 2.26 & 25 & 840238 / 152739 \\
PE\_2 & 33.1 & 15 & 70.8 & 1.12 & 0.020 & 10 (19) & 1.37 & 19 & 783028 / 904370 \\
PE\_3 & 35.2 & 21 & 81.9 & 1.15 & 0.064 & 11 (19) & 2.20 & 15 & 962891 / 137395 \\
PEG\_1 & 24.0 & 23 & 64.0 & 1.00 & 0.052 & 20 (29) & 0.75 & 74 & 148731 / 135693 \\
PEG\_2 & 24.1 & 15 & 57.2 & 1.05 & 0.022 & 20 (29) & 1.97 & 112 & 798838 / 316927 \\
PEG\_3 & 23.5 & 14 & 61.7 & 1.04 & 0.055 & 20 (29) & 0.59 & 142 & 200187 / 631302 \\
PLLA\_1 & 15.2 & 18 & 37.2 & 1.04 & 0.010 & 18 (24) & 0.03 & 96 & 686928993 / 561359054 \\
PLLA\_2 & 16.6 & 22 & 39.3 & 1.11 & 0.010 & 19 (24) & 0.05 & 58 & 412260453 / 328430378 \\
PLLA\_3 & 16.8 & 26 & 41.8 & 1.12 & 0.018 & 18 (24) & 0.03 & 33 & 967838819 / 25364483 \\
aPS\_1 & 14.7 & 12 & 35.7 & 1.30 & 0.021 & 25 (27) & 0.18 & 212 & 120191275 / 979826601 \\
aPS\_2 & 14.8 & 14 & 31.4 & 1.34 & 0.027 & 24 (27) & 0.03 & 54 & 211721699 / 422597568 \\
aPS\_3 & 14.1 & 13 & 29.9 & 1.28 & 0.046 & 24 (27) & 0.20 & 239 & 796234216 / 992835374 \\
sPVC\_1 & 22.5 & 23 & 58.8 & 1.34 & 0.033 & 19 (32) & 2.77 & 35 & 414243003 / 405550636 \\
sPVC\_2 & 21.1 & 17 & 51.2 & 1.28 & 0.024 & 20 (32) & 3.69 & 103 & 604339280 / 438801734 \\
sPVC\_3 & 20.9 & 20 & 49.7 & 1.33 & 0.027 & 19 (32) & 1.78 & 184 & 492680472 / 585758416 \\
PEEK\_1 & 25.8 & 22 & 64.6 & 1.27 & 0.021 & 16 (22) & 0.13 & 166 & 768489 / 766602 \\
PEEK\_2 & 25.4 & 19 & 67.8 & 1.22 & 0.015 & 16 (22) & 0.03 & 146 & 791429 / 898589 \\
PEEK\_3 & 26.6 & 24 & 64.4 & 1.29 & 0.042 & 16 (22) & 0.07 & 135 & 973247 / 55037 \\
PSU\_1 & 22.3 & 23 & 55.6 & 1.20 & 0.009 & 18 (23) & 0.05 & 57 & 5396 / 836097 \\
PSU\_2 & 18.0 & 18 & 34.9 & 1.07 & 0.010 & 18 (23) & 0.13 & 50 & 804277 / 685700 \\
PSU\_3 & 19.5 & 16 & 43.8 & 1.12 & 0.009 & 18 (23) & 0.17 & 117 & 675306 / 454398 \\
\end{longtable}
\endgroup
\normalsize

\section{Agent Ablation}
\label{sec:si_recovery}

\paragraph{Arms.} Recovery is scored as completion: the failed stage must be accepted and the run must yield a gate-passing property. PTFE is not graded against experiment. PTFE\_noAI and PTFE\_AI ran from the same validated plan (OPLS-AA, DP~50, 10 chains, 100~K\,ns$^{-1}$ cooling) with identical seeds and deterministic remedies; without the agent, an issue no remedy addresses ends the run. RadonPy's stock workflow is an external reference, not an ablation arm, and has no spatial-homogeneity check.

\paragraph{Failure and agent decisions.} Both melts failed only the density-homogeneity check (0.193 without and 0.185 with the agent, limit 0.11), and the no-agent run ended. The agent ended its run, diagnosing the check's independent-atom counting floor as too low when 76\% of the mass is fluorine bonded in CF$_2$ groups, since every mixing indicator passed and no plan revision changes the measured quantity. After the authors revised the check, the agent run completed. When its $T_g$ fits disagreed by 72.4~K, the agent accepted the stage with $T_g$ withheld so that the mechanical stage could run.

\paragraph{Check revision.} The original check subtracts a counting floor that assumes independently placed atoms; all 21 campaign melts pass it (largest signal 0.071). The revised melt check compares voxel density maps averaged over the two halves of the melt hold and counts only structure present in both (limit 0.11). Regraded without new simulation, PTFE\_AI gives 0.027 and PTFE\_noAI 0.000, so the no-agent melt would also pass. PTFE\_AI's melt was first accepted under an intermediate form (0.033, limit 0.08) replaced the same day. No result in Section~S5 depends on the revision.

\paragraph{Outcome.} Completion required the authors' correction; the agent's autonomous contribution was the thermal-stage acceptance, which the engine routes only to the agent.

\bibliography{reference}